\documentclass{llncs}

\usepackage{graphicx}
\usepackage{amssymb}
\usepackage{amsmath}

\newcommand{\setcomp}[2]{\ensuremath{\{#1\ |\ #2\}}}
\newcommand{\set}[1]{\ensuremath{\{#1\}}}

\newcommand{\bbsum}[2]{\ensuremath{\overset{\!\!\!#1}{\underset{#2}{\sum}\ }}}

\newcommand{\bbmin}[2]{\ensuremath{\overset{#1}{\underset{#2}{\min}\ }}}

\usepackage{pgf,tikz}
\usetikzlibrary{arrows,automata}

\usepackage{multirow}

\newcommand{\myproof}{\noindent {\bf Proof:\ \ }}
\newcommand{\myqed}{\mbox{$\Box$}}

\begin{document}
\sloppy

\mainmatter

\title{The \textsc{RegularGcc} Matrix Constraint}
\titlerunning{The \textsc{RegularGcc} Matrix Constraint}
\author{Ronald de Haan\inst{1}\inst{2} \and
Nina Narodytska\inst{2}\inst{3} \and
Toby Walsh\inst{2}\inst{3} 
\\
\url{Ronald.de\_Haan@mailbox.tu-dresden.de, \{nina.narodytska,toby.walsh\}@nicta.com.au}}
\authorrunning{Ronald de Haan, Nina Narodytska, Toby Walsh}
\institute{Technische Universit\"at Dresden\\
\and
NICTA, Australia
\and 
University of New South Wales}

\toctitle{The \textsc{RegularGcc} matrix constraint}
\tocauthor{Ronald de Haan, Nina Narodytska, Toby Walsh}
\maketitle


\begin{abstract}
We study propagation of the \textsc{RegularGcc} global
constraint. This ensures that each row of a matrix of
decision variables satisfies a \textsc{Regular} constraint,
and each column satisfies a \textsc{Gcc} constraint.
On the negative side, we prove that propagation is NP-hard even under some
strong restrictions (e.g. just 3 values, just 4 states
in the automaton, or just 5 columns to the matrix).
On the positive side, we identify two cases where propagation is fixed
parameter tractable. In addition, we show how to improve
propagation over a simple decomposition into separate 
\textsc{Regular} and \textsc{Gcc} constraints by identifying
some necessary but insufficient conditions for a solution.
We enforce these conditions with some
additional weighted row automata. Experimental
results demonstrate the potential of these methods
on some standard benchmark problems.
\end{abstract}

\section{Introduction}
Global constraints can be used to model and reason
about commonly found substructures. 
Many such models contain matrices of decision variables 
\cite{Flener:2001th,Flener:2002ux,ReginG04}.
Matrix constraints are global constraints that apply to such
matrices \cite{Katsirelos:wm}. For example, the \textsc{RegularGcc}
matrix constraint can be used to model rostering problems. 
It ensures each row of the matrix satisfies a \textsc{Regular}
constraint (representing the shift rules) and each column
satisfies a \textsc{Gcc} constraint (representing the required
capacities for each shift). 
We prove here that propagating the \textsc{RegularGcc} constraint is costly, even under very severe restrictions. Therefore, as in \cite{Beldiceanu:2010wk},
we look for partial methods that only enforce a limited level of consistency. These methods are based on necessary conditions that improve propagation over the decomposition into separate \textsc{Regular} constraints on the rows and separate \textsc{Gcc} constraints on the columns. 
These necessary conditions depend on extracting several string properties from the rows. We enforce these necessary conditions by constraining the rows with additional automaton constraints. Unfortunately, when the number of columns increases, these automata increase in size quite drastically. By using weighted automata, we show that we can limit the increase in size. Finally, we 
show that this approach can be used in a more general
setting where we have a matrix with \textsc{multicostRegular} and \textsc{Gcc} constraints.


\section{Intractable cases}\label{sec:complexity}
We first prove that propagating the \textsc{RegularGcc}  matrix
constraint
is intractable even under strong conditions. More precisely,
we show that enforcing bound consistency (BC) is NP-hard. 
This justifies why we later look for partial propagation
methods based on some necessary (but not sufficient) conditions. 

\begin{theorem}\label{thm:bcregulargcc}
Enforcing BC on \textsc{RegularGcc} is \textsc{NP}-hard, already for \textsc{Regular} constraints given by a DFA of size $4$, \textsc{Gcc} constraints specifying only an upper bound on the number of occurrences of one particular value, and just $3$ values.
\end{theorem}
\myproof
Reduction from 3-SAT. Let $\varphi = \gamma_1 \wedge \cdots \wedge \gamma_C$ be a Boolean formula in CNF on propositional variables $p_1,\ldots,p_R$. We construct an $R \times C$ matrix $\mathcal{M}$ of decision variables taking their values from $\set{-1,0,1}$, where each row $1 \leq r \leq R$ corresponds to a propositional variable $p_r$ and each column $1 \leq c \leq C$ corresponds to a clause $\gamma_c$.

To initialize the domain of variables in the matrix, we do the following for each clause $\gamma_i = l^i_1 \vee l^i_2 \vee l^i_3$. We set $\mathcal{M}_{r,i} = 0$ for all propositions $p_r$ not occurring in $\gamma_i$. For $j \in \set{1,2,3}$ we set $\mathcal{M}_{i,k} \in \set{0,1}$ if $l^i_j = p_k$ and we set $\mathcal{M}_{i,k} \in \set{0,-1}$ if $l^i_j = \neg p_k$.

On each column we put the \textsc{Gcc} constraint that states that the value $0$ occurs at most $R-1$ times. On each row we put the \textsc{Regular} constraint that states that besides $0$'s either only $1$'s or only $-1$'s occur.

We show that this instance of \textsc{RegularGcc} has a solution iff $\varphi$ is satisfiable.

$(\Rightarrow)$ We create a satisfying assignment $I$ for $\varphi$ as follows. For each $p_r$, if in row $r$ occurs at least one $1$, we let $I(p_r) = \top$, otherwise we let $I(p_r) = \bot$ (the choice of $I(p_r)$ when only $0$'s occur in row $r$ is arbitrary). Since in each column $c$ there occur only $R-1$ many $0$'s, we know that there exists some $p_i$ for which $\mathcal{M}_{i,c} \neq 0$ and thus $I(l^c_i) = \top$. Therefore $I \models \gamma_c$.

$(\Leftarrow)$ Let $I$ be an assignment satisfying $\varphi$. We can instantiate $\mathcal{M}$ as follows. For each clause $\gamma_c = l^c_1 \vee l^c_2 \vee l^c_3$, for $j \in \set{1,2,3}$ we do the following.
If $l^c_j = p_k$ and $I(p_k) = \top$, we let $\mathcal{M}_{k,c} = 1$. If $l^c_j = \neg p_k$ and $I(p_k) = \bot$, we let $\mathcal{M}_{k,c} = -1$. Otherwise, we let $\mathcal{M}_{k,c} = 0$.
Since $I$ is functional each \textsc{Regular} constraint on the rows is satisfied. Also, since at least one literal is satisfied in each clause, each column contains at least one value that is not $0$, so the \textsc{Gcc} constraints are satisfied.

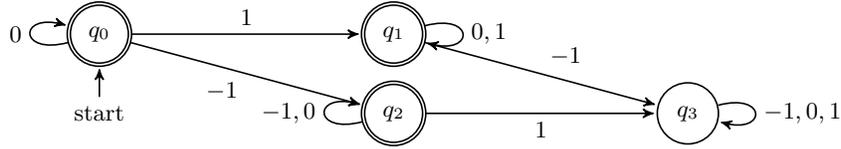
\begin{figure}[]
\begin{center}
\begin{tikzpicture}[->,>=stealth',shorten >=1pt,auto,node distance=1.1cm,semithick]
  \tikzstyle{every state}=[]
  \node[state,accepting,initial below] (q0) {$q_{0}$};
  \node[state,accepting] (q1) [right of=q0, xshift=80pt] {$q_{1}$};
  \node[state,accepting] (q2) [right of=q0, xshift=80pt, yshift=-30pt] {$q_{2}$};
  \node[state] (q3) [right of=q2, xshift=80pt] {$q_{3}$};
  \path (q0) edge [loop left] node {$0$} (q0);
  \path (q0) edge [] node {$1$} (q1);
  \path (q0) edge [swap] node {$-1$} (q2);
  \path (q1) edge [loop right] node {$0,1$} (q1);
  \path (q1) edge [] node {$-1$} (q3);
  \path (q2) edge [loop left] node {$-1,0$} (q2);
  \path (q2) edge [swap] node {$1$} (q3);
  \path (q3) edge [loop right] node {$-1,0,1$} (q3);
\end{tikzpicture}
\end{center}
\vspace{-20pt}
\caption{Automaton $\mathcal{A}$.}
\label{fig:automaton}
\end{figure}

Automaton $\mathcal{A}$ in Fig.~\ref{fig:automaton} witnesses that this \textsc{Regular} constraint can be enforced by a DFA of size $4$.
Note that this proof also works for any other restriction on \textsc{Regular} constraints that can enforce that, for two given values, in any word at most one of these values occurs.
\myqed

In fact, since we only bound the number of one particular value in the \textsc{Gcc} constraint, the above proof also works for the \textsc{RegularAmong} constraint.

A common type of \textsc{Regular} constraint in a \textsc{RegularGcc}
matrix constraint
is a \textsc{Stretch} constraint. This constrains the length of 
any stretch of values (e.g. there are 
at most 3 night shifts in a row) and
the possible transitions (e.g. a night shift can only be followed
by a day off). Unfortunately, even this special case is intractable
to propagate. 

\begin{theorem}
Enforcing BC on \textsc{StretchGcc} is \textsc{NP}-hard, already for just $3$ values.
\end{theorem}
\myproof
Reduction from the Exact Cover problem. We are given $F = \set{S_1,\ldots,S_n}$ with $\bigcup_{i}S_i = U$. We ask if there is some subset $C \subseteq F$ with $\bigcup_{c \in C}c = U$ and $c \cap c' = \emptyset$ for all distinct $c,c' \in C$. W.l.o.g. we assume $U$ contains the integers $1$ to $|U|$.

We construct a $|F| \times |U|$ matrix $\mathcal{M}$, of decision variables taking their value in $\set{-1,0,1}$. For each row $1 \leq r \leq |F|$ and each value $1 \leq i \leq |U|$ we do the following. If $i \in S_r$, we let $\mathcal{M}_{r,i} \in \set{0,1}$. If $i \not\in S_r$, we let $\mathcal{M}_{r,i} \in \set{-1,0}$.

On each column we put the \textsc{Gcc} constraint that states that the value $1$ occurs exactly once. On each row we put the \textsc{Stretch} constraint stating that each stretch of $0$'s must have a length of at least $|U|$.

We show that this instance of \textsc{StretchGcc} has a solution iff there exists an exact cover.

$(\Rightarrow)$ Take a solution for our instance. We let $C$ be the set of all $U_r$ for which row $r$ in the solution contains only $-1$'s and $1$'s. Obviously $C \subseteq F$. In order to show that $\bigcup_{c \in C} = U$, it suffices to show that $U \subseteq \bigcup_{c \in C}$. Take an arbitrary $i \in U$. Since our solution contains at least one $1$ in each column, we know there is some $c \in C$ such that $i \in c$. We also show that all distinct $c,c' \in C$ are disjoint. Take arbitrary $c,c' \in C$ such that $c \neq c'$. Assume $j \in c \cap c'$. This means that column $j$ in the solution would contain two $1$'s, which contradicts the \textsc{Gcc} constraints on the columns.

$(\Leftarrow)$ Let $C \subseteq F$ be an exact cover. We fill $\mathcal{M}$ as follows. For each row $1 \leq r \leq |F|$, we do the following. If $S_r \in C$, fill the row with $-1$'s and $1$'s (this can be done only in one way). Otherwise, fill row $r$ with only $0$'s. Obviously, the \textsc{Stretch} constraints on the rows are satisfied. Also, since $C$ is an exact cover, we know that for each $1 \leq i \leq |U|$ there is exactly one row $r$ such that $\mathcal{M}_{r,i} = 1$. Thus the \textsc{Gcc} constraints on the columns are satisfied.
\myqed

This proof also works for the \textsc{Regular} constraint accepting only words that contain either only $0$'s or only $-1$'s and $1$'s, and therefore for any restriction on $\textsc{Regular}$ constraints that can enforce permitted (or forbidden) words of length two (such as the meta constraint \textsc{Slide}).

\begin{corollary}
Enforcing BC on \textsc{SlideGcc} is \textsc{NP}-hard, already for \textsc{Slide} constraints based on constraints of arity $2$ and just $3$ values.
\end{corollary}

Another common type of \textsc{Regular} constraint in a \textsc{RegularGcc} 
matrix constraint is
a \textsc{Sequence} constraint. This limits the number of values
of a particular type that occur in each sequence (e.g. at most
3 shifts in every 7 can be night shifts). This case is 
unfortunately also intractable to propagate. We
prove that both enforcing domain consistency (DC) and
enforcing bound consistency (BC) are NP-hard even if the matrix
has just a few columns.

\begin{theorem}\label{thm:dcsequencegcc}
Enforcing DC on \textsc{SequenceGcc} is \textsc{NP}-hard, already for just $5$ columns.
\end{theorem}
\myproof
Reduction from the 3D Matching problem. The proof is inspired by \cite{Crama95}. Given are three pair-wise disjoint sets $W$, $Z$, $Y$ of equal size $q$ and a set $M \subseteq W \times Z \times Y$, $|M| = m$. The question is if there exists $M' \subseteq M$ such that $|M'| = q$ and no two different elements of $M'$ agree in any coordinate.

Assume $M = \set{s_1,\ldots,s_m}$. We create a $m \times 5$ matrix $\mathcal{M}$ of decision variables taking their value in $\set{0,t,w_1,\ldots,w_q,z_1,\ldots,z_q,y_1,\ldots,y_q}$. For each $(w_{i},z_{i},y_{i}) = s_i$ we let $\mathcal{M}_{i,1} \in \set{0,w_i}$, $\mathcal{M}_{i,2} \in \set{0,t}$, $\mathcal{M}_{i,3} \in \set{0,z_i}$, $\mathcal{M}_{i,4} \in \set{0,t}$, and $\mathcal{M}_{i,5} \in \set{0,y_i}$.

We constrain each row $i$ with the constraint \textsc{Sequence}$(\mathcal{M}_i, 1, 2, 2, \set{0})$, stating that in each sequence of length $2$, at least one $0$ occurs. On columns $1$ (resp. $3$ and $5$) we put the \textsc{Gcc} constraint stating that each value in $W$ (resp. $Z$ and $Y$) occurs at least once, and that at least $m-q$ many $0$'s occur. On columns $2$ and $4$ we put the \textsc{Gcc} constraint stating that at least $q$ many $t$'s occur, and at least $m-q$ many $0$'s.

We show that this instance of \textsc{SequenceGcc} has a solution iff there exists a 3D matching.

$(\Rightarrow)$ Take a solution for \textsc{SequenceGcc}. We know column $2$ contains exactly $m-q$ many $t$'s, and $q$ many $0$'s. For each occurrence of a $t$ in column $2$ at row $i$, columns $1$ and $3$ contain a $0$ at row $i$ (by the \textsc{Sequence} constraint). Then, by the \textsc{Gcc} constraint, for all rows $j$ where column $2$ contains a $0$, columns $1$ and $3$ contain a non-$0$ at row $j$, and thus (by \textsc{Sequence}) column $4$ contains a $0$ at row $j$. By a similar argument, we know that in the remaining rows column $4$ contains $t$'s. Continuing this argument for column $5$, we know that in the solution there are $q$ many rows taking values $(w_{i},0,z_{i},0,y_{i})$ and $m-q$ rows taking values $(0,t,0,t,0)$. By the \textsc{Gcc} constraints, we know that each value $w \in W$ occurs exactly once, as well as each value $z \in Z$ and each $y \in Y$. Since the possible values were chosen by taking elements from $M$, we know that $M' = \setcomp{s_i}{\mathcal{M}_{i} \neq (0,t,0,t,0)}$ is a 3D matching.

$(\Leftarrow)$ Let $M' \subseteq M$ be a 3D matching. We can fill $\mathcal{M}$ as follows. For each $(w_i,z_i,y_i) = s_i \in M'$, we let $\mathcal{M}_{i} = (w_{i},0,z_{i},0,y_{i})$. For each $s_i \in M \backslash M'$ we let $\mathcal{M}_{i} = (0,t,0,t,0)$. Obviously each row satisfies the \textsc{Sequence} constraint. Since $|M'| = q$ and each value $w \in W$ occurs exactly once in the first coordinate of $M'$ (and similarly for values $z \in Z$ and the second coordinate, and $y \in Y$ and the third coordinate), we have that each column satisfies the corresponding \textsc{Gcc} constraint.
\myqed

Note that the \textsc{Sequence} constraints on all rows are the same, but the \textsc{Gcc} constraints on the columns differ.

\begin{theorem}\label{thm:bcsequencegcc}
Enforcing BC on \textsc{SequenceGcc} is \textsc{NP}-hard, already for just $5$ columns.
\end{theorem}
\myproof
The proof is similar to the proof of Theorem~\ref{thm:dcsequencegcc}, and also inspired by \cite{Crama95}.

Let $cw_{i}$ (resp. $cz_{i}$, $cy_{i}$) be the number of occurrences of the value $w_{i}$ (resp. $z_{i}$, $y_{i}$) in $M$. For each value $w_{i}$ (resp. $z_{i}$, $y_{i}$) we create $cw_{i} -1$ (resp. $cz_{i} -1$, $cy_{i} -1$) clones of it. We now define the total order on values as $U = [-w_{q}^{1},\ldots,-w_{q}^{cw_{q}-1},\ldots,-w_{1}^{1},\ldots,-w_{1}^{cw_{1}-1},
-z_{q}^{1},\ldots,-z_{q}^{cz_{q}-1},\ldots,-z_{1}^{1},\ldots,$ $-z_{1}^{cz_{1}-1},
-y_{q}^{1},\ldots,-y_{q}^{cy_{q}-1},\ldots,-y_{1}^{1},\ldots,-y_{1}^{cy_{1}-1},
0,t,y_{1},\ldots,y_{q},z_{1},\ldots,z_{q},w_{1},$ $\ldots,w_{q} ]$.

We create the matrix $\mathcal{M}$ in a similar fashion as in the proof for Theorem~\ref{thm:dcsequencegcc}, with the difference that for each $(w_{i},z_{i},y_{i}) = s_i$ we let $\mathcal{M}_{i,1} \in [-w^{1}_{i},\ldots,-w^{cw_{i}-1}_{i}, \ldots, w_{i}]$, $\mathcal{M}_{i,3} \in [-z^{1}_{i},\ldots,-z^{cz_{i}-1}_{i}, \ldots, z_{i}]$, $\mathcal{M}_{i,5} \in [-y^{1}_{i},\ldots,-y^{cy_{i}-1}_{i}, \ldots, y_{i}]$, 
and $\mathcal{M}_{i,2} \in [0,t]$ and $\mathcal{M}_{i,4} \in [0,t]$.

We adapt the constraint on rows to \textsc{Sequence}$(\mathcal{M}_i, 1, 2, 2, [-w_{q}^{1},0])$, stating that in each sequence of length $2$, at least one value in $[-w_{q}^{1},0]$ occurs. On columns $1$ (resp. $3$ and $5$) we replace the \textsc{Gcc} constraint with one stating that each value in $\set{-w_{q}^{1},\ldots,-w_{q}^{cw_{q}-1},\ldots,-w_{1}^{1},\ldots,-w_{1}^{cw_{1}-1}} \cup W$ (resp. $\set{-z_{q}^{1},\ldots,-z_{q}^{cz_{q}-1},\ldots,-z_{1}^{1},\ldots,-z_{1}^{cz_{1}-1}} \cup Z$ and $\set{-y_{q}^{1},\ldots,-y_{q}^{cy_{q}-1},\ldots,-y_{1}^{1},\ldots,-y_{1}^{cy_{1}-1}} \cup Y$) occurs at least once. We do not change the \textsc{Gcc} constraints on columns $2$ and $4$.

We show that this instance of \textsc{SequenceGcc} has a solution iff there exists a 3D matching.

$(\Rightarrow)$ By reasoning similarly to the proof of Theorem~\ref{thm:dcsequencegcc} (replacing `$0$' with `a value in $[-w_{q}^{1},0]$' when reasoning about columns $1$, $3$ and $5$), we know that in the solution there are $q$ many rows taking values $(w_{i},0,z_{i},0,y_{i})$, possibly containing clones, and $m-q$ rows taking values $(n,t,n',t,n'')$ where $n,n',n''$ are either $0$ or some clone $w'_{i}$, $z'_{i}$, or $y'_{i}$ (respectively).

Furthermore, by the \textsc{Gcc} constraint on the odd columns, we know that each value in $\set{-w_{q}^{1},\ldots,-w_{q}^{cw_{q}-1},w_{q}}$ must occur exactly once. Since these values occur only in the domains of $\mathcal{M}_{i,1}$ for the $cw_{q}$ many $s_i \in M$ that contain $w_{q}$, we know that each of these $\mathcal{M}_{i,1}$ must take a value in $\set{-w_{q}^{1},\ldots,-w_{q}^{cw_{q}-1},w_{q}}$.

Then, by the \textsc{Gcc} constraint on the odd columns, we know that each value in $\set{-w_{q-1}^{1},\ldots,-w_{q-1}^{cw_{q-1}-1},w_{q-1}}$ must occur exactly once. These values occur only in the domains of $\mathcal{M}_{i,1}$ for the $s_i \in M$ that contain $w_{q-1}$ or $w_{q}$. However, the $\mathcal{M}_{i,1}$ for the $s_i \in M$ that contain $w_{q}$ must take values in $\set{-w_{q}^{1},\ldots,-w_{q}^{cw_{q}-1},w_{q}}$. Therefore, the $\mathcal{M}_{i,1}$ for the $s_i \in M$ that contain $w_{q-1}$ must take a value in $\set{-w_{q-1}^{1},\ldots,-w_{q-1}^{cw_{q-1}-1},w_{q-1}}$.

Repeating this argument recursively until reaching the value $t$, we can restrict the effective domain of the odd positions of $\mathcal{M}_{i}$ for $s_{i} = (w_{i},z_{i},y_{i}) \in M$ to $\mathcal{M}_{i,1} \in \set{-w^{1}_{i}, \ldots, -w^{cw_{i}-1}_{i}, w_{i}}$, $\mathcal{M}_{i,3} \in \set{-z^{1}_{i}, \ldots, -z^{cz_{i}-1}_{i}, z_{i}}$, and $\mathcal{M}_{i,5} \in \set{-y^{1}_{i}, \ldots, -y^{cy_{i}-1}_{i}, y_{i}}$.

Therefore, we know that each row $\mathcal{M}_i$ for $s_{i} = (w_{i},z_{i},y_{i}) \in M$ either has the values $(w_{i},0,z_{i},0,y_{i})$ or the values $(w'_{i},t,z'_{i},t,y'_{i})$ for some clones $w'_{i},z'_{i},y'_{i}$ of $w_{i},z_{i},y_{i}$. Now, by the \textsc{Gcc} constraints, we know that each value $w \in W$ occurs exactly once, as well as each value $z \in Z$ and each $y \in Y$. Since the possible values were chosen by taking elements from $M$, we know that $M' = \setcomp{s_i}{\mathcal{M}_{i} = (w,0,z,0,y), w \in W, z \in Z, y \in Y}$ is a 3D matching.

$(\Leftarrow)$ Let $M' \subseteq M$ be a 3D matching. We can fill $\mathcal{M}$ as follows. For each $(w_i,z_i,y_i) = s_i \in M'$, we let $\mathcal{M}_{i} = (w_{i},0,z_{i},0,y_{i})$. For each $(w_i,z_i,y_i) = s_i \in M \backslash M'$ we let $\mathcal{M}_{i} = (w'_{i},t,z'_{i},t,y'_{i})$ for some clones $w'_{i},z'_{i},y'_{i}$ of $w_{i},z_{i},y_{i}$ that have not been used before in the process of filling $\mathcal{M}$. We know there are enough different clones for this procedure. It is easy to verify that this instantiation of $\mathcal{M}$ satisfies all the constraints.
\myqed

\section{Fixed parameter tractable cases}

We have seen that propagating
the \textsc{RegularGcc} matrix constraint is \textsc{NP}-hard even under the strong restriction that either the number of values or the number of columns is bounded.
However, if we consider \textsc{Regular$^2$} and we bound the number of columns and the number of states in the row and column automata
we at last have a case in which 
propagation is polynomial.

\begin{theorem}
Enforcing DC on \textsc{Regular$^2$} is fixed parameter tractable in $k = C\cdot|Q|\cdot(\log |Q'|)$, where $C$ is the number of columns, $|Q|$ is the size of the row automata, and $|Q'|$ is the size of the column automata.
\end{theorem}
\myproof
We assume w.l.o.g. that all row constraints are the same, and all column constraints are the same. We can encode the matrix constraint on a $R \times C$ matrix $\mathcal{M}$ in a single DFA on the matrix stretched out to a single sequence of variables $\mathcal{M}_{1,1},\ldots,\mathcal{M}_{1,C},\ldots,\mathcal{M}_{R,1},\ldots,\mathcal{M}_{R,C}$. The state set of the automaton is $Q \times Q'^{|C|}$. In each state, the automaton keeps track of the current state $q' \in Q'$ for each column $c$, as well as the current state $q \in Q$ in the current row. The size of the automaton is $\mathcal{O}(|Q| \cdot 2^{C(\log |Q|)})$. Enforcing DC on a \textsc{Regular} constraint takes time polynomial in the size of the automaton, so our algorithm runs in fixed parameter tractable time.
\myqed

We also get tractability if we bound the number of rows and the size
of the automata. 
 
\begin{theorem}
Enforcing DC on \textsc{RegularGcc} is fixed parameter tractable in $k = r(\log Q)$ where $r$ is the number of rows and $Q$ the maximum number of states in any row automaton.
\end{theorem}
\myproof
This follows directly from Observation~2 in \cite{Katsirelos:wm}, and the fact that \textsc{Gcc} over a sequence with fixed size can be encoded in a DFA with polynomially many states.
\myqed

On the other hand, just bounding the number of rows is not
enough to give tractability. 

\begin{theorem}\label{thm:regulargccw2hard}
Enforcing BC on \textsc{RegularGcc} is W[2]-hard in $k = R$ the number of rows, even with just $2$ values.
\end{theorem}
\myproof
This proof is similar to the proof of Theorem~3 in \cite{Katsirelos:wm}. We reduce from \textsc{$p$-Hitting-Set}. Let $\mathcal{H} = (V,E)$ a hypergraph, where $V = \set{v_1,\ldots,v_{|V|}}$ and $E = \set{e_1,\ldots,e_{|E|}}$. We ask if there is a hitting set $S \subseteq V$ in $\mathcal{H}$ of cardinality $k$. 

We construct an instance $\mathcal{M}$ of \textsc{RegularGcc} with $|V| + |E|$ columns and $k$ rows on the alphabet $\set{0,1}$.
The \textsc{Regular} constraint accepts $|V|$ different words $w^{1},\ldots,w^{|V|}$ of length $|V| + |E|$. For any word $w^{v}$, the $v$th value is $1$, and the remainder of the first $|V|$ values is $0$. Also, for any word $w^{v}$ and any $1 \leq j \leq |E|$, the $j$th value of $w^{v}$ is $1$ if $v \in e_j$, and is $0$ otherwise.
The \textsc{Gcc} constraints we put on the columns are as follows. In the first $|V|$ columns we require exactly one $1$. In the remaining columns, we require at least one $1$.

In this reduction, each row corresponds to one vertex that is being chosen for inclusion in the hitting set, and each column after the first $|V|$ to one hyperedge.

The \textsc{Gcc} constraints on the first $|V|$ columns ensure that no vertex is chosen in two rows. The constraints on the last $|E|$ columns ensure that each hyperedge contains a vertex chosen for the hitting set. If there is a hyperedge all vertices of which are not included in the hitting set, the column corresponding to this hyperedge will contain $0$'s only, violating the \textsc{Gcc} constraint of that column.

We show that there exists a hitting set $S$ in $\mathcal{H}$ of cardinality $k$ iff the \textsc{RegularGcc} matrix constraint has a solution.

$(\Rightarrow)$ Assume there exists a hitting set $S$ in $\mathcal{H}$ of size $k$. We construct an assignment to \textsc{RegularGcc} by matching one vertex $v \in S$ with each row (in any manner). If row $i$ is matched to vertex $v$, we assign word $w^{v}$ to row $i$. The fact that $S$ contains $k$ different vertices $v$ ensures that $k$ different words $w^v$ are used. Also, since $S$ is a hitting set, we know that for each hyperedge $e_{j}$ there is at least one $v \in e_{j} \cap S$, and so each of the last $|E|$ columns contains at least one $1$. Thus the \textsc{Gcc} constraints are satisfied.

$(\Leftarrow)$ Suppose the \textsc{RegularGcc} matrix constraint has a solution. We construct a hitting set $S$ by taking all $v$ such that $w^{v}$ is a row in the solution. By the \textsc{Gcc} constraints we know that the solution contains $k$ different words $w^v$, so $S$ is of size $k$. Now, to derive the contrary, assume there exists a hyperedge $e_{j} \in E$ such that $S \cap e_{j} = \emptyset$. Then the column corresponding to $e_{j}$ (column $|V| + j$) contains only $0$'s. This violates \textsc{Gcc} on this column, which is a contradiction.
\myqed

Note that the \textsc{Gcc} constraints in the above proof can be expressed with \textsc{Regular} constraints of bounded size as well, which gives us the following corollary.

\begin{corollary}
Enforcing BC on the \textsc{Regular$^{2}$} matrix constraint is W[2]-hard in $k$ the number of rows, even with just $2$ values.
\end{corollary}

Another special case that is intractable is when
we repace the \textsc{Gcc} constraint on the columns
with a simpler sum constraint. 

\begin{theorem}
Enforcing BC on the \textsc{RegularSum} matrix constraint is W[2]-hard in $k$ the number of rows, even with just $3$ values.
\end{theorem}
\myproof (Sketch)
The proof is similar to the proof of Theorem~\ref{thm:regulargccw2hard} and the proof of Theorem~3 in \cite{Katsirelos:wm}. We reduce from \textsc{$p$-Hitting-Set} and construct a matrix constraint as in the proof of Theorem~\ref{thm:regulargccw2hard}, with the following differences. The first $|V|$ columns we fill with $-1$'s instead of $1$'s. We can then replace the \textsc{Gcc} constraints on these columns with the \textsc{Sum} constraint requiring a sum of at least $-1$. The \textsc{Gcc} constraints on the last $|E|$ columns can be replaced with the \textsc{Sum} constraint requiring a sum of at least $1$. The arguments in the proof of Theorem~\ref{thm:regulargccw2hard} now hold for this instance of \textsc{RegularSum}.
\myqed

Note that this result is strictly stronger than the W[1]-hardness proof of enforcing BC on \textsc{RegularSum} in \cite{Katsirelos:wm}.

\section{Some necessary conditions}\label{sec:necessary}

Motivated by these rather negative complexity results, we investigate how to improve propagation over a simple decomposition into separate \textsc{Regular} and \textsc{Gcc} constraints by means of deriving necessary conditions based on string properties. In fact, we will show how to extend the method of \cite{Beldiceanu:2010wk} to the (decomposed) setting of \textsc{multicostRegular} constraints on the rows and \textsc{Gcc} constraints on the columns. This method is based on a double counting argument. Using automata constraints we extract several string properties from the rows. For these string properties, we derive lower and upper bounds based on the \textsc{Gcc} constraints on the columns. This allows us to derive necessary constraints relating the bounds to the corresponding string properties.

We start with some preliminary definitions needed for our exposition. The \textsc{multicostRegular} global constraint \cite{Menana:2009up} is defined as follows. Given a sequence $X = (x_1,x_2,\ldots,x_n)$ of finite domain decision variables and a deterministic finite automaton $\mathcal{A} = (Q,V,\Delta,s,F)$, the constraint \textsc{Regular}$(X,\mathcal{A})$ holds iff $X$ is a word of length $n$ over $V$ accepted by DFA $\mathcal{A}$. Given a vector $Z = (z^0,\ldots,z^{R})$ of bounded variables and $c = (c^r_{q,v})^{r\in [0\ldots R]}_{q\in Q,v \in V}$ a family of assignment cost matrices, \textsc{multicostRegular}$(X,Z,\mathcal{A},c)$ holds iff \textsc{Regular}$(X,\mathcal{A})$ holds and for an accepting run $q_0q_1\ldots q_n$ of $\mathcal{A}$ on the instantiation $(v_0,\ldots,v_n)$ of $X$ we have that $\sum_{0 \leq i < n}{c^r_{q_{i},v_{i+1}} = z^r}$ for all $0 \leq r \leq R$.

For any two DFAs $\mathcal{A}_1 = (Q_1,V,\Delta_1,s_1,F_1)$ and $\mathcal{A}_2 = (Q_2,V,\Delta_2,s_2,F_2)$, with corresponding $c^1$ and $c^2$ cost matrices over resources $\mathcal{R} = \set{r^0,\ldots,r^R}$, we define the product automaton $\mathcal{A}_1 \times \mathcal{A}_2 = (Q_1 \times Q_2, V, \Delta, (s_1,s_2), F_1 \times F_2)$ and product cost matrix $c = c^1 \times c^2$ as follows.
\[ \Delta((q_1,q_2),v) = (\Delta_1(q_1,v),\Delta_2(q_2,v)) \]
\[ c^{r}_{(q_1,q_2),v} = c^{1,r}_{q_1,v} + c^{2,r}_{q_2,v} \quad \mbox{for $0 \leq r \leq R$}\]
In other words, when taking the product of two weighted automata, we take the usual cross product of the underlying automata, and add the cost matrices.

We show how to extract relevant string properties using \textsc{multicostRegular} constraints on the rows. In the following, we let $v \in V$ denote a value that the decision variables can take, we let $\hat{v}, \hat{w}_{i} \subseteq V$ (for indices $i \in \mathbb{N}$) denote a subset of these values, we let $\neg\hat{v}$ denote $V \backslash \hat{v}$, and we let $Z$ be a set of bounded variables representing the calculated weights. We also define the concatenation $\hat{w}_1 \cdot \ldots \cdot \hat{w}_m$ of several $\hat{w}_i$ as the set $\setcomp{w_1\cdots w_m \in V^{m}}{w_i \in \hat{w}_i, 1 \leq i \leq m}$.

\begin{figure}
\vspace{-10pt}
\begin{center}
\begin{tikzpicture}[->,>=stealth',shorten >=1pt,auto,node distance=1.6cm,semithick]
  \tikzstyle{every state}=[]
  \node[state,initial left,accepting] (q0) {$q_0$};
  \node[state,accepting] (q1) [right of=q0, xshift=50pt] {$q_1$};
  \path (q0) edge [loop right] node {$\neg\hat{v}, 0$} (q0);
  \path (q0) edge [bend left=15] node {$\hat{v}, 1$} (q1);
  \path (q1) edge [loop right] node {$\hat{v}, 0$} (q1);
  \path (q1) edge [bend left=15, -] node {$\neg\hat{v}, 0$} (q0);
\end{tikzpicture}
\end{center}
\vspace{-20pt}
\caption{Automaton $\mathcal{A}^{\hat{v}}_1$. Transitions are marked with cost $c^{r}$.}
\label{fig:automaton1}
\end{figure}
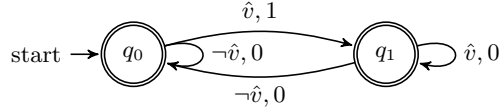

To extract the number of uninterrupted stretches of elements from $\hat{v}$ in $X$ using a resource variable $z^{r} \in Z$, we can use the weighted DFA $\mathcal{A}^{\hat{v}}_1$ (Figure~\ref{fig:automaton1}), where transitions are marked with the symbol and the cost $c^r$ they consume. For any word $X$, we have that \textsc{multicostRegular}$(X,Z,\mathcal{A}^{\hat{v}}_1,c)$ holds for $z^{r}$ the number of stretches of symbols in $\hat{v}$ that occur in $X$.

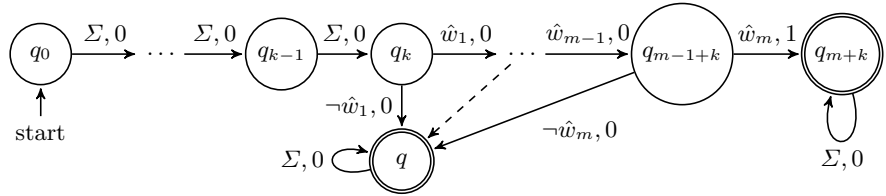
\begin{figure}
\vspace{-10pt}
\begin{center}
\begin{tikzpicture}[->,>=stealth',shorten >=1pt,auto,node distance=1.6cm,semithick]
  \tikzstyle{every state}=[]
  \node[state,initial below] (q0) {$q_{0}$};
  \node[] (qa) [right of=q0] {$\ldots$};
  \node[state] (qi) [right of=qa] {$q_{k-1}$};
  \node[state] (q0plusi) [right of=qi] {$q_{k}$};
  \node[] (qb) [right of=q0plusi] {$\ldots$};
  \node[state] (qmmin1plusi) [right of=qb, xshift=15pt] {$q_{m-1+k}$};
  \node[state,accepting] (qmplusi) [right of=qmmin1plusi, xshift=15pt] {$q_{m+k}$};
  \node[state,accepting] (qc) [below of=q0plusi, yshift=5pt] {$q$};
  \path (q0) edge [] node {$\Sigma,0$} (qa);
  \path (qa) edge [] node {$\Sigma,0$} (qi);
  \path (qi) edge [] node {$\Sigma,0$} (q0plusi);
  \path (q0plusi) edge [] node {$\hat{w}_1,0$} (qb);
  \path (qb) edge [] node {$\hat{w}_{m-1},0$} (qmmin1plusi);
  \path (qb) edge [dashed] node {} (qc);
  \path (qmmin1plusi) edge [] node {$\hat{w}_m,1$} (qmplusi);
  \path (q0plusi) edge [swap] node {$\neg\hat{w}_1,0$} (qc);
  \path (qmmin1plusi) edge [] node {$\neg\hat{w}_m,0$} (qc);
  \path (qc) edge [loop left] node {$\Sigma, 0$} (qc);
  \path (qmplusi) edge [loop below] node {$\Sigma, 0$} (qmplusi);
\end{tikzpicture}
\end{center}
\vspace{-15pt}
\caption{Automaton $\mathcal{A}^{k,\overline{\hat{w}}}_2$. Transitions are marked with cost $c^{r}$.}
\label{fig:automaton2}
\end{figure}

To extract whether a word $\overline{w} \in \overline{\hat{w}}$ occurs in $X$ starting at position $k$ using a resource variable $z_{k}^{r}$, we can use the weighted DFA $\mathcal{A}^{k,\overline{\hat{w}}}_2$ (Figure~\ref{fig:automaton2}) with parameter $k \in \mathbb{N}$, where transitions are marked with the symbol and the cost $c^r$ they consume. For any word $X$, we have that \textsc{multicostRegular}$(X,Z,\mathcal{A}^{k,\overline{\hat{w}}}_2,c)$ sets $z_{k}^{r}$ to true if and only if some word $\overline{w} \in \overline{\hat{w}}$ occurs in $X$ starting at position $k$. To extract the total number of occurrences of words $\overline{w} \in \overline{\hat{w}}$ (starting any position) in $X$, we take the sum of the values of the variables $z_{k}^{r}$ (for $1 \leq k \leq n$) that represent whether a suitable word $\overline{w}$ occurs in $X$ starting at position $k$.

To extract the minimum and maximum length of stretches, we can simulate counters using weights. Let $\mathcal{A}$ be a DFA annotated with counters $d = (d_1,\ldots,d_m)$, taking their values from $\set{0,\ldots,n-1}$. We can construct a DFA $\mathcal{A}'$ of size less than or equal to $n^m \cdot |\mathcal{A}|$, together with a cost matrix $c$ for resources $r^1,\ldots,r^m$ such that for any word $w$
there exists an accepting run for $w$ on $\mathcal{A}$
where the counters have final values $(v_1,\ldots,v_m)$
if and only if
there exists an accepting run for $w$ on $\mathcal{A}'$
for $(z^1,\ldots,z^m) = (v_1,\ldots,v_m)$.
This can be done straightforwardly by choosing $Q \times \set{0,\ldots,n-1}^m$ as state set for $\mathcal{A}'$, and choosing transitions $\Delta$ corresponding to the update formulae for the counters. Now $c$ can be chosen to mimic the changes in counter values over transitions. The automaton $\mathcal{A}'$ can possibly be reduced in size by removing unreachable states or minimizing it using other methods.

We can transform any given automaton $\mathcal{A}$ to extract the minimum and maximum length of a stretch of symbols from $\hat{v}$ occurring in $\mathcal{A}$ on $X$ as follows. We annotate $\mathcal{A}$ with counters that represent \textit{stretchminlen}($\hat{v},n$) and \textit{stretchmaxlen}($\hat{v},n$), as described in \cite{Beldiceanu:2010wk}. Then we transform this annotated automaton, as described above, into an automaton $\mathcal{A}'$ with resource variables $z^{\hat{v}}_{min}$ and $z^{\hat{v}}_{max}$ whose values (respectively) represent the minimum and maximum length of stretches of symbols in $\hat{v}$ occurring in $X$.

The above automata, extracting the different string properties from rows, can be combined with each other and with other automata by using the product operation. By defining zero cost matrices for all resources not used explicitly in a given automaton, we can extract several different string properties simultaneously with one weighted product automaton.

In this more general decomposed setting with \textsc{multicostRegular} constraints on the rows, a tractable option for propagation is the algorithm based on a Lagrangian relaxation of the Resource Constrained Shortest Path Problem (RCSPP) from \cite{Menana:2009up}. Using weighted automata to extract string properties has several advantages. The size of the automata is relatively low. The automata used to extract the number of stretch occurrences are even of constant size. A weighted automaton used to extract a string property is never larger than the unfolding of an (unweighted) automaton annotated with counters used to extract the same string property. Also, the use of weighted automata allows us to express several other constraints with small automata. For instance, \textsc{Gcc} constraints on the rows can be expressed by a weighted automaton with a single state. In fact, \textsc{Gcc} constraints can be expressed using additional weights on other automaton constraints already posed on the rows.

Using the above string properties, we can derive necessary conditions that exploit the matrix structure. Consider the following CSP, similar to the one sketched in \cite{Beldiceanu:2010wk}. Given three positive integers $R$, $K$, and $V$, we have an $R \times K$ matrix $\mathcal{M}$ of decision variables with domain $\set{0,1,\ldots,V-1}$, and a $V \times K$ matrix $\mathcal{M}^{\#}$ of cardinality variables with domain $\set{0,1,\ldots,R}$. Each row $r$, for $0 \leq r < R$, of $\mathcal{M}$ is subject to a \textsc{multicostRegular} constraint. For simplicity, we assume that each row is subject to the same constraint. Each column $k$, for $0 \leq k < K$, of $\mathcal{M}$ is subject to a \textsc{Gcc} constraint that restricts the number of occurrences of the values according to column $k$ of $\mathcal{M}^{\#}$. Let $\#^{v}_{k}$ denote the number of occurrences of value $v$, for $0 \leq v < V$, in column $k$ of $\mathcal{M}$, that is, the cardinality variable in row $v$ and column $k$ of $\mathcal{M}^{\#}$. For any $\hat{v} \subseteq V$, we let $\#^{\hat{v}}_{k}$ denote $\sum_{v \in \hat{v}}(\#^{v}_{k})$.

In order to constrain the number of occurrences of words,
we define the bounds $lw_k(\overline{\hat{w}})$ and $uw_k(\overline{\hat{w}})$ on the number of occurrences of words in $\overline{\hat{w}}$ starting at column $k$, based on the \textsc{Gcc} constraints on the columns, as follows:
\begin{eqnarray}
lw_k(\overline{\hat{w}}) &=& \max \left ( \left ( \hspace{2pt} \bbsum{|\overline{\hat{w}}|-1}{j=0} \#^{\hat{w}_j}_{k+j} \right ) - (|\overline{\hat{w}}| - 1) \cdot R, 0 \right )
\label{eq:lw}
\\
uw_k(\overline{\hat{w}}) &=& \bbmin{|\overline{\hat{w}}| - 1}{j=0} \left ( \#^{\hat{w}_j}_{k+j} \right )
\label{eq:uw}
\end{eqnarray}
Note that definitions (\ref{eq:lw}) and (\ref{eq:uw}) are exactly the same as in \cite{Beldiceanu:2010wk}. The lower bound (\ref{eq:lw}) is the worst-case intersection of all column value occurrences. The upper bound (\ref{eq:uw}) is justified by the fact that a word cannot occur more often than its minimally occurring letter. We now get the following necessary conditions for each $0 \leq k < K$:\\
\begin{minipage}{0.5\textwidth}
\begin{equation}
lw_k(\overline{\hat{w}}) \leq \bbsum{R-1}{r=0} z^{\overline{\hat{w}}}_{r,k}
\label{eq:wordocc1}
\end{equation}
\end{minipage}
\begin{minipage}{0.5\textwidth}
\begin{equation}
uw_k(\overline{\hat{w}}) \geq \bbsum{R-1}{r=0} z^{\overline{\hat{w}}}_{r,k}
\label{eq:wordocc2}
\end{equation}
\end{minipage}\\[5pt]
where $z^{\overline{\hat{w}}}_{r,k}$ denotes the resource variable representing whether a word in $\overline{\hat{w}}$ occurs in row $r$ starting at column $k$. Since we extracted the number of word occurrences for each starting position $k$, we can directly relate the bounds derived from the column constraints with the number of word occurrences per starting position. This results in constraints (\ref{eq:wordocc1}) and (\ref{eq:wordocc2}) potentially leading to more propagation than their counterparts in \cite{Beldiceanu:2010wk}. This is illustrated in Example~\ref{ex:morepropagation}.
Note that the constraints from \cite{Beldiceanu:2010wk} on words occurring as a prefix or as a suffix correspond to particular cases of constraints (\ref{eq:wordocc1}) and (\ref{eq:wordocc2}).

\begin{example}\label{ex:morepropagation}
Consider the scenario concerning a partially instantiated $5 \times 5$ matrix in Figure~\ref{fig:example}, which could occur as a node in the search tree. Let $\overline{\hat{w}} = \set{2}\set{2}$. In this scenario $lw_k(\overline{\hat{w}})$ are variables. Also, $z^{\overline{\hat{w}}}_{r}$ is a variable such that $z^{\overline{\hat{w}}}_{r} = \sum_{k=0}^{K-1} z^{\overline{\hat{w}}}_{r,k}$. In this scenario, the bounds of the variables $z^{\overline{\hat{w}}}_{r,k}$ can be automatically derived by the row automata. By using equation (\ref{eq:wordocc1}), we can directly detect unsatisfiability in this case, since $lw_1(\overline{\hat{w}}) \in [3,5]$ and thus $lw_1(\overline{\hat{w}}) \not\leq ( \sum^{R-1}_{r=0} z^{\overline{\hat{w}}}_{r,1}) \in [0,2]$. Consider the counterpart of equation (\ref{eq:wordocc1}) from \cite{Beldiceanu:2010wk}: $\sum^{K-|\overline{\hat{w}}|}_{k=0}lw_k(\overline{\hat{w}}) \leq \sum^{R-1}_{r=0} z^{\overline{\hat{w}}}_{r}$. Using this constraint, unsatisfiability cannot directly be detected in this particular case.
\end{example}

\begin{figure}[]
\vspace{-20pt}
\caption{Example search tree node.}
\label{fig:example}
\hspace{20pt}\begin{minipage}{90pt}
$\mathcal{M}$:
\begin{tabular}{||p{10pt}|p{10pt}|p{10pt}|p{10pt}|p{10pt}||}
\hline
&&&&\\
\hline
&&&&\\
\hline
\hspace{2.5pt}1&&\hspace{2.5pt}1&&\\
\hline
\hspace{2.5pt}1&&\hspace{2.5pt}1&&\\
\hline
\hspace{2.5pt}1&&\hspace{2.5pt}1&&\\
\hline
\end{tabular}
\end{minipage}
\begin{minipage}{270pt}
\begin{center}
\begin{tabular}{p{80pt} p{70pt} p{80pt}}
$lw_0(\overline{\hat{w}}) \in [0,2]$&$z^{\overline{\hat{w}}}_{0,1} \in [0,1]$&$z^{\overline{\hat{w}}}_{0} \in [0,1]$\\
$lw_1(\overline{\hat{w}}) \in [3,5]$&$z^{\overline{\hat{w}}}_{1,1} \in [0,1]$&$z^{\overline{\hat{w}}}_{1} \in [0,1]$\\
$lw_2(\overline{\hat{w}}) \in [0,2]$&$z^{\overline{\hat{w}}}_{2,1} = 0$&$z^{\overline{\hat{w}}}_{2} \in [0,1]$\\
$lw_3(\overline{\hat{w}}) \in [0,5]$&$z^{\overline{\hat{w}}}_{3,1} = 0$&$z^{\overline{\hat{w}}}_{3} \in [0,1]$\\
$lw_4(\overline{\hat{w}}) = 0$&$z^{\overline{\hat{w}}}_{4,1} = 0$&$z^{\overline{\hat{w}}}_{4} \in [0,1]$\\
\end{tabular}
\end{center}
\end{minipage}
\end{figure}

Take note of the following case, where $\overline{\hat{w}} = \hat{v}$ for some $\hat{v} \subseteq V$. In this case, for each $0 \leq k < K$ the constraints (\ref{eq:lw}) and (\ref{eq:uw}) and the constraints (\ref{eq:wordocc1}) and (\ref{eq:wordocc2}) simplify to, respectively:\\
\begin{minipage}{0.5\textwidth}
\begin{equation}
lw_{k}(\hat{v}) = uw_{k}(\hat{v}) = \#^{\hat{v}}_{k}
\end{equation}
\end{minipage}
\begin{minipage}{0.5\textwidth}
\begin{equation}
\#^{\hat{v}}_{k} = \bbsum{R-1}{r=0} z^{\hat{v}}_{r,k}
\end{equation}
\end{minipage}


In order to constrain the number of occurrences of stretches,
we define the bounds $ls^{+}_{k}$ and $us^{+}_{k}$ (referring to the number of uninterrupted stretches of variables from $\hat{v}$ that start in column $k$) and the bounds $ls^{-}_{k}$ and $us^{-}_{k}$ (referring to the number of uninterrupted stretches of variables from $\hat{v}$ that end in column~$k$), based on the \textsc{Gcc} constrains as follows:
\begin{eqnarray}
ls^{+}_{k} &=& \max(0, \#^{\hat{v}}_{k} - \#^{\hat{v}}_{k-1}) \label{eq:lsplus} \\
us^{+}_{k} &=& \#^{\hat{v}}_{k} - \max(0, \#^{\hat{v}}_{k-1} + \#^{\hat{v}}_{k} - R) \label{eq:usplus} \\
ls^{-}_{k} &=& \max(0, \#^{\hat{v}}_{k} - \#^{\hat{v}}_{k+1}) \label{eq:lsmin} \\
us^{-}_{k} &=& \#^{\hat{v}}_{k} - \max(0, \#^{\hat{v}}_{k+1} + \#^{\hat{v}}_{k} - R) \label{eq:usmin}
\end{eqnarray}

Definitions (\ref{eq:lsplus}) through (\ref{eq:usmin}) are exactly the same as in \cite{Beldiceanu:2010wk}. The lower bound (\ref{eq:lsplus}) is the difference between the number of occurrences of values $\hat{v}$ in column $k$ minus the number of occurrences of $\hat{v}$ in column $k-1$, if positive. If the total number of occurrences of values $\hat{v}$ on column $k$ and on column $k-1$ are strictly greater than the number of rows $R$, then there must be at least $\#^{\hat{v}}_{k-1}+\#^{\hat{v}}_{k}-R$ stretches of values $\hat{v}$ that cover both columns. This minimum intersection gives us the upper bound (\ref{eq:usplus}). Bounds (\ref{eq:lsmin}) and (\ref{eq:usmin}) are derived similarly. We now get the following necessary conditions:\\
\begin{minipage}{0.5\textwidth}
\begin{eqnarray}
\bbsum{K-1}{k=0}ls^{+}_{k}(\hat{v}) &\leq& \bbsum{R-1}{r=0} z^{\hat{v}}_{r} \\
\bbsum{K-1}{k=0}us^{+}_{k}(\hat{v}) &\geq& \bbsum{R-1}{r=0} z^{\hat{v}}_{r} 
\end{eqnarray}
\end{minipage}
\begin{minipage}{0.5\textwidth}
\begin{eqnarray}
\bbsum{K-1}{k=0}ls^{-}_{k}(\hat{v}) &\leq& \bbsum{R-1}{r=0} z^{\hat{v}}_{r} \\
\bbsum{K-1}{k=0}us^{-}_{k}(\hat{v}) &\geq& \bbsum{R-1}{r=0} z^{\hat{v}}_{r}
\end{eqnarray}
\end{minipage}\\[5pt]
where $z^{\hat{v}}_{r}$ denotes the variable corresponding to the resource that represents the number of uninterrupted sequences of symbols in $\hat{v}$ occurring in row $r$.


In order to constrain the minimum and maximum length of a stretch,
using the minimum and maximum length ($z^{\hat{v}}_{min}$ and $z^{\hat{v}}_{max}$, respectively) of uninterrupted sequences of symbols in $\hat{v}$ occurring in any row, we get the following necessary conditions for each $0 \leq k < K$:\\[5pt]
\begin{minipage}{0.5\textwidth}
\begin{equation}
\#^{\hat{v}}_{k} \geq \bbsum{k}{j = \max(0,k-z^{\hat{v}}_{min}+1)} ls^{+}_{j}(\hat{v}) \label{eq:strlen1}
\end{equation}
\end{minipage}
\begin{minipage}{0.5\textwidth}
\begin{equation}
\#^{\hat{v}}_{k} \geq \bbsum{\min(K-1,k+z^{\hat{v}}_{min}-1)}{j=k} ls^{-}_{j}(\hat{v}) \label{eq:strlen2}
\end{equation}
\end{minipage}\\[10pt]

Constraints (\ref{eq:strlen1}) and (\ref{eq:strlen2}) are justified by the fact that stretches starting resp. ending at the considered columns $j$ must overlap column $k$. Also, for each $0 \leq k < K - z^{\hat{v}}_{max}$ we get the necessary condition:
\begin{eqnarray}
ls^{+}_{k}(\hat{v}) + \bbsum{z^{\hat{v}}_{max}}{j=z^{\hat{v}}_{min}} \#^{\hat{v}}_{k+j} - (z^{\hat{v}}_{max} - z^{\hat{v}}_{min}+1) \cdot R &\leq& 0 \label{eq:strlen3}
\end{eqnarray}
and for each $z^{\hat{v}}_{max} \leq k < K$ the necessary condition:
\begin{eqnarray}
ls^{-}_{k}(\hat{v}) + \bbsum{z^{\hat{v}}_{max}}{j=z^{\hat{v}}_{min}} \#^{\hat{v}}_{k-j} - (z^{\hat{v}}_{max} - z^{\hat{v}}_{min}+1) \cdot R &\leq& 0 \label{eq:strlen4}
\end{eqnarray}
The justification behind constraint (\ref{eq:strlen3}) is that for a stretch of values $\hat{v}$ beginning at column $k$, there must be a value not in $\hat{v}$ in some column $j$, for $k + z^{\hat{v}}_{min} \leq j \leq k + z^{\hat{v}}_{max}$. Constraint (\ref{eq:strlen4}) is justified similarly.

\section{Evaluation}\label{sec:evaluation}

To evaluate our method, we used NSPLib \cite{Vanhoucke:2005vt}, a library of benchmark instances of the nurse scheduling problem (NSP). This is a particular rostering problem. For $N$ the number of nurses, $D$ the number of days in the scheduling horizon, and $S$ the number of shifts, the objective is to construct a $N \times D$ matrix of values in the integer interval $[1, S]$, where value $S$ represents the off-duty shift.

In instance files, there are hard coverage constraints and soft preference constraints. We only consider the hard coverage constraints. These give for each day $d$ and shift $s$ the lower bound on the number of nurses that must be assigned to shift $s$ on day $d$. These constraints can be modelled by \textsc{Gcc} constraints on the columns. We considered instance files for $N \times 7$ rosters with $N \in \set{25, 50, 75, 100}$.

Case files provide hard constraints on the rows. For each shift $s$, there are lower and upper bounds on the number of occurrences of $s$ in any row. There are also lower and upper bounds on the cumulative number of occurrences of the working shifts $1,\ldots,S-1$ in any row. These two types of constraints can be modelled by \textsc{Gcc} constraints on the rows. For each shift $s$, there are also lower and upper bounds on the length of any stretch of value $s$ in any row. Finally, there are lower and upper bounds on the length of any stretch of working shifts $1,\ldots,S-1$ in any row. These two types of constraints can be modelled by \textsc{stretch\_path} and \textsc{stretch\_path\_partition} constraints on the rows, respectively. By translating these row constraints to automata, we get that the NSP benchmark problems as described above correspond to the \textsc{RegularGcc} pattern studied in this paper.

In order to compare the effect of the necessary conditions in the settings of both weighted and unweighted automata, we implemented the row constraints (both for the constraints from the case files and for extracting string properties) using weighted finite automata as well as regular (unweighted) finite automata. For the setting of unweighted automata, we translated the case constraints specified for each shift and for the total set of working shifts as a single \textsc{Regular} constraint on each row (by taking the corresponding minimised product DFA). For each string property that we extract from the rows, we used automata annotated with counters (as described in \cite{Beldiceanu:2010wk}), unfolded into a DFA, expressed as a decomposition into ternary constraints \cite{qwcp06} allowing us to extract the counter values. The methods used in \cite{Beldiceanu:2010wk} for automata annotated with counters are not implemented in the free major constraint programming libraries and solvers.

For the setting of weighted automata. We translated the case constraints for each shift and for the total set of working shifts as a single \textsc{multicostRegular} constraint on each row (by taking the corresponding product automaton). For each string property that we extract from the rows, we posed a single \textsc{multicostRegular} constraint defined by the corresponding weighted automaton as described in Section~\ref{sec:necessary}.

In order to compare the two settings fairly, we posed the constraints defined by automata in a similar pattern, i.e., we take the products of corresponding automata in the two settings. One advantage that the setting of weighted automata possesses, is that taking the product of particular automata results in a relatively small increase in the automaton size, not nearly as explosive as the size increase in the corresponding unweighted product automaton. In order to improve propagation, we were able to pose the weighted automata extracting the number of stretches of different shifts from the rows as the product of the corresponding automaton with (a copy of) the automata specifying the constraints on the number of shift occurrences from the case file. In the unweighted setting this is completely intractable, since the size of the product DFA corresponding to the automata annotated with counters gets too large.

In both settings, we implemented necessary constraints based on the following string properties:
\begin{itemize}
  \item for each shift, lower and upper bounds on the number of its occurrences,
  \item for each shift, lower and upper bounds on the number and length of its stretches,
  \item each word of length at most 2 that consists of one single shift.
\end{itemize}
In the setting of weighted automata, the necessary constraints are derived as described in Section~\ref{sec:necessary}. In the setting of unweighted automata, the necessary constraints are derived as in \cite{Beldiceanu:2010wk}.

The objective of our experiments is to measure the impact in runtime and backtracks for the different settings. The experiments were run under Choco 2.1.1 on a 2.27 GHz Intel Xeon with a 4GB RAM. All runs were allocated 3 CPU minutes. For each case and nurse count $N$, we used instances 1-270.

In the experiments we used a labelling procedure that selects variables with the smallest domain, with a row-wise order as tie-breaker, and selects the smallest value. We used a \textsc{LexChain} constraint for symmetry breaking. We used the implementation of the \textsc{multicostRegular} constraint available in Choco.

Table~\ref{tab:benchmarkresults} summarises the running of the instances for the different settings (the setting of weighted automata with cross products (CWA) and without extra cross products (WA), and the setting of unweighted automata (UA)), for Cases 7 and 8. Each row first indicates the number of known instances of some satisfiability status for a given case and nurse count $N$, and then the performance of each setting to the first solution, namely the number of instances decided to be of that status without timing out, as well as the average runtime (in seconds) and the average number of backtracks for all instances on which none of the settings timed out. Numbers in boldface indicate best performance in a row.

The benchmark results in Table~\ref{tab:benchmarkresults} show that WA and CWA were able to solve significantly more instances compared to the method using unweighted automata, both for satisfiable and unsatisfiable instances.
Further, CWA improved the performance for most of the benchmarks in terms of number of backtracks and runtime, compared to WA.
Notably, the UA method solved only 4 out of 156 known unsatisfiable instances while CWA and UA solved all of these benchmarks.
This shows that using weighted automata together with necessary constraints leads to significantly more pruning than using unweighted automata with similar necessary constraints. 
For the majority of solved unsatisfiable instances, WA and CWA detected unsatisfiability at the root of the search tree.
This is not visible in the table, because the shown runtimes and number of backtracks are based on instances solved by all methods.
Note that these benchmarking results are not directly comparable to the results in \cite{Beldiceanu:2010wk}, since these results were obtained under a different experimental set-up (e.g. a different search strategy was used).

Overall, the results indicate that the use of weighted automata to solve rostering problems shows potential.
A combination of weighted automata and necessary constraints dramatically increase propagation compared to using unweighted automata.
Our results on unsatisfiable instances suggest that such a combination can be very useful in finding optimum solutions for rostering problems.
Another advantage of our approach is that it can be easily implemented in open-source constraint solvers.

\begin{table}[]
\vspace{-15pt}
\caption{NSPLib benchmark results.}
\begin{center}
\begin{scriptsize}
\begin{tabular}{|r|r|r|r||r|r|r||r|r|r||r|r|r|}
\hline
\multicolumn{4}{|c|}{}&\multicolumn{3}{c|}{WA}&\multicolumn{3}{c|}{CWA}&\multicolumn{3}{c|}{UA}\\\hline
Case&N&Status&Known&\#Inst&Time&\#Bktk&\#Inst&Time&\#Bktk&\#Inst&Time&\#Bktk\\
\hline
7&25&sat&129&122&23.8&1866&\textbf{123}&21.9&\textbf{1509}&103&\textbf{21.1}&2400\\
&&unsat&30&\textbf{30}&0&0&\textbf{30}&0&0&0&0&0\\
7&50&sat&60&58&\textbf{16.6}&\textbf{693}&\textbf{60}&19.5&708&34&20.0&1227\\
&&unsat&31&\textbf{31}&\textbf{0.1}&\textbf{0}&\textbf{31}&0.3&\textbf{0}&1&0.2&\textbf{0}\\
7&75&sat&29&25&\textbf{22.0}&742&\textbf{27}&25.6&\textbf{737}&17&22.1&929\\
&&unsat&38&\textbf{38}&0&0&\textbf{38}&0&0&0&0&0\\
7&100&sat&34&29&30.9&1733&\textbf{34}&\textbf{29.8}&\textbf{1437}&13&38.5&2196\\
&&unsat&19&\textbf{19}&0.2&\textbf{0}&\textbf{19}&\textbf{0.2}&\textbf{0}&1&0.3&\textbf{0}\\
\hline
8&25&sat&138&131&11.5&776&\textbf{133}&10.6&\textbf{646}&114&\textbf{9.1}&1123\\
&&unsat&6&\textbf{6}&0&0&\textbf{6}&0&0&0&0&0\\
8&50&sat&90&83&13.1&606&\textbf{88}&\textbf{9.4}&\textbf{294}&71&15.0&1512\\
&&unsat&8&\textbf{8}&\textbf{0.1}&\textbf{0}&\textbf{8}&0.1&\textbf{0}&1&0.2&\textbf{0}\\
8&75&sat&61&58&13.3&412&\textbf{62}&\textbf{10.4}&\textbf{233}&45&12.6&505\\
&&unsat&19&\textbf{19}&0&0&\textbf{19}&0&0&0&0&0\\
8&100&sat&65&60&17.9&308&\textbf{65}&\textbf{13.4}&\textbf{143}&45&16.3&439\\
&&unsat&5&\textbf{5}&\textbf{0.1}&\textbf{0}&\textbf{5}&0.1&\textbf{0}&1&0.3&\textbf{0}\\
\hline
\end{tabular}
\end{scriptsize}
\end{center}
\label{tab:benchmarkresults}
\end{table}
%

\vspace{-40pt}
\section{Conclusions}
We studied the propagation of the \textsc{RegularGcc} matrix constraint.
We showed that propagation is NP-hard, even under some strong restrictions,
and also showed two cases in which propagation is fixed parameter tractable.
Additionally, we showed how to improve propagation over a decomposition
into separate \textsc{Regular} constraints on the rows and \textsc{Gcc} constraints on the columns
by identifying some necessary but insufficient conditions.
We showed how the use of weighted automata for the row constraints can be beneficial.
Experimental results on nurse scheduling problems demonstrate the potential for this method.

\pagebreak
\bibliographystyle{splncs03}
\bibliography{globalconstraints}

\end{document}